\documentclass[sigconf]{acmart}

\setcopyright{none}
\renewcommand\footnotetextcopyrightpermission[1]{}
\acmYear{2026}
\acmConference[MiLeTS '26]{The 12th Mining and Learning from Time Series Workshop}{August 2026}{Jeju, Republic of Korea}
\acmBooktitle{The 12th Mining and Learning from Time Series Workshop (MiLeTS '26), held in conjunction with KDD 2026}

\usepackage{booktabs}
\usepackage{multirow}
\usepackage{graphicx}
\usepackage{subcaption}
\usepackage{xcolor}
\usepackage{listings}
\usepackage{tikz}
\usepackage{tabularx}
\usetikzlibrary{shapes.geometric, arrows.meta, positioning, calc, fit, backgrounds}

\title[LeNEPA: No-Augmentation Next-Latent Prediction for Time Series]{\texorpdfstring{LeNEPA: No-Augmentation Next-Latent Prediction\\for Time-Series Representation Learning}{LeNEPA: No-Augmentation Next-Latent Prediction for Time-Series Representation Learning}}

\author{Alexander Chemeris}
\affiliation{
  \institution{Langotime}
  \country{South Africa}
}
\email{alex@langotime.ai}

\author{Ming Jin}
\affiliation{
  \institution{Griffith University}
  \country{Australia}
}

\author{Randall Balestriero}
\affiliation{
  \institution{Brown University}
  \country{United States}
}

\begin{document}

\begin{abstract}
Time series are central to modern data mining applications, from industrial telemetry and server metrics to finance and physiology, yet time-series self-supervised learning often depends on view and augmentation choices that encode domain-specific invariances.
We study a narrower practical question than fully tuned benchmark performance: how does an SSL recipe behave when its method-specific configuration is reused unchanged after the pretraining signal family changes?
We frame this as a fixed-recipe stress test rather than as a comparison against optimally tuned methods.
We introduce Latent Euclidean Next-Embedding Prediction Architecture (LeNEPA), a no-augmentation next-latent-token prediction objective with a causal backbone.
LeNEPA replaces the stop-gradient/EMA stabilization used by vanilla NEPA with SIGReg-based isotropy regularization and computes the predictive loss in a lightweight projected space that is discarded for evaluation.
We compare LeNEPA with an ECG-tuned JEPA recipe under a fixed-horizon frozen-probe protocol on PTB-XL and Diag, a synthetic diagnostic corpus generated with Aionoscope.
Both methods are retrained independently on each dataset, but their method-specific recipes are kept unchanged: LeNEPA keeps the same no-augmentation objective configuration, while JEPA keeps the same ECG-tuned masking recipe.
In this protocol, the ECG-tuned JEPA recipe is strong in-domain on PTB-XL but weaker when reused unchanged on Diag, whereas LeNEPA preserves useful frozen-probe gains on both datasets.
Learning curves suggest faster early representation acquisition under this fixed-horizon protocol: LeNEPA reaches $80\%$ of its final AUROC/AUPRC gain after $2$--$5$k updates, compared with $5$--$10$k updates for the faster JEPA readout.
As a separate external frozen-encoder check, a CauKer-pretrained LeNEPA variant reaches $77.65\%$ mean UCR-128 Random-Forest accuracy in a single-seed, best-checkpoint run---within $1.16$ points of Mantis and within $0.24$ points of MOMENT ($77.89\%$).
Overall, the results support no-augmentation latent prediction as a useful candidate recipe for low-retuning time-series SSL.
Source code and artifacts are available at \url{https://github.com/langotime/lenepa-milets-2026}.
\end{abstract}

\ccsdesc[500]{Computing methodologies~Machine learning}
\ccsdesc[300]{Mathematics of computing~Time series analysis}

\keywords{time-series self-supervised learning, time-series representation learning, joint embedding predictive architectures, augmentation-free learning, frozen-feature time-series encoders}

\maketitle

\section{Introduction}
\label{sec:introduction}

Time series are the natural interface to an evolving world: they record how systems change, spanning server metrics, sensor streams, industrial and telecommunication telemetry, finance, and physiology. As analytical systems and AI agents increasingly monitor, diagnose, and intervene in real time, portable pretraining recipes become useful infrastructure for downstream classification, anomaly understanding, root-cause analysis, and clinical diagnosis support. Despite rapid progress for images and text, however, building time-series encoders that require little per-dataset engineering remains difficult: pipelines are still task-specific, and pretrained backbones are far from standardized. A practical SSL recipe should therefore expose compact sequence embeddings while reducing the amount of view, augmentation, and probe engineering needed for each new data source.

Many successful self-supervised objectives (contrastive learning~\citep{chen2020simclr,he2020moco}, masked reconstruction~\citep{he2021mae}, and modern JEPA variants~\citep{assran2023ijepa,bardes2024vjepa,balestriero2025lejepa}) rely on carefully designed views that preserve task-specific semantics. In vision, this design space is comparatively well understood, yet representation quality can still be sensitive to view engineering: in DINO, removing multi-crop views reduces ImageNet linear-probe top-1 from $76.1\%$ to $72.5\%$ (ViT-S/16), and varying the multi-crop scale split shifts k-NN top-1 from $65.6\%$ to $69.8\%$~\citep{caron2021dino}; this dependence is even stronger in SimCLR and BYOL~\citep{chen2020simclr,grill2020byol}. In time series---where what constitutes a ``safe'' transformation depends on data type, sampling, and downstream semantics~\citep{yue2022ts2vec,eldele2021tstcc,woo2022cost,tonekaboni2021tnc}---this sensitivity is amplified: an augmentation study reports that augmentation choice alone can create up to $32$ accuracy points on a fault-detection dataset, and that the best augmentation reaches F1 $0.6975$ vs.\ $0.4811$ with no pretraining~\citep{liu2024tsaugselect}. In our experiments, small changes in augmentation recipes for time-series JEPA lead to large swings in representation quality (Section~\ref{sec:ablations_diagnostics}).

These considerations expose a trade-off.
Augmentation-based JEPA-style methods~\citep{assran2023ijepa} are well studied and can produce strong representations (e.g., ECG\_JEPA for PTB-XL~\citep{weimann2024jepaecg}), but their performance depends critically on augmentation choices.
\textbf{When a view recipe is reused outside the domain it was designed for, augmentation engineering can become a practical bottleneck for reusable time-series SSL recipes.}

\paragraph{Contributions}
\begin{itemize}
  \item We propose LeNEPA, a \emph{no-augmentation next-embedding prediction architecture} for frozen-backbone time-series representation learning, adapting SIGReg and Guillotine Regularization to autoregressive latent prediction to remove stop-gradient/EMA stabilization from the tested recipe.

  \item We evaluate fixed-recipe pretraining reuse across real ECG data and Diag, a synthetic diagnostic corpus generated with the Aionoscope library~\citep{chemeris2026aionoscope}, separating recipe portability from checkpoint transfer: LeNEPA preserves useful frozen-probe gains when reused unchanged, while an ECG-tuned JEPA view recipe is strong in-domain but weaker when its fixed recipe is reused on a structurally different signal family.

  \item We compare a frozen CauKer-pretrained LeNEPA encoder on UCR-128 against Mantis, NuTime, and MOMENT Random-Forest protocol anchors, using no handcrafted augmentations.
\end{itemize}

\begin{figure*}[t]
\centering
\resizebox{0.82\linewidth}{!}{%
\begin{tikzpicture}[
    font=\sffamily\footnotesize, thick,
    enc/.style   ={draw, rounded corners, minimum width=2.9cm, minimum height=0.8cm, align=center, fill=green!12},
    proj/.style  ={draw, trapezium, trapezium angle=65, minimum width=1cm, minimum height=0.75cm, align=center, fill=orange!18},
    loss/.style  ={draw, rounded corners, minimum width=2.4cm, minimum height=0.95cm, align=center, fill=blue!10},
    io/.style    ={align=center},
    note/.style  ={align=center, font=\sffamily\scriptsize, text=black!75},
    arrow/.style ={-{Latex[length=2mm]}, line width=1pt},
    tap/.style   ={-{Latex[length=2mm]}, line width=0.9pt, densely dashed, black!55}
  ]
    \node[io] (input) at (0,0) {Input series\\$x\in\mathbb{R}^{C\times L}$};
    \node[enc, above=0.7cm of input] (emb) {Conv patch\\embedding $\tau_\theta$};
    \node[enc, minimum height=2.25cm, above=0.55cm of emb] (vit) {Causal ViT $f_\phi$};
    \draw[arrow] (input) -- (emb);
    \draw[arrow] (emb) -- node[right=1pt, font=\scriptsize]{$z_{1:T}$} (vit);

    \node[proj, rotate=-90] (pproj) at ($(vit.north east)+(4.0,0.5)$) {Projector\\$h_\psi$};
    \node[proj, rotate=-90] (tproj) at ($(emb.east)+(4.0,0)$) {Projector\\$h_\psi$};

    \coordinate (predmid) at ($(emb.east)!0.5!(tproj.south)$);
    \draw[arrow] (vit.north) -- (vit.north |- pproj.south) -- (pproj.south);
    \node[above, font=\scriptsize] at (predmid |- pproj.south) {prediction $\hat z_{t}$};
    \node[below, font=\scriptsize, text=red!60!black] at (predmid |- pproj.south) {no stop-grad};
    \draw[arrow] (emb.east) -- node[above, font=\scriptsize]{target $z_{t+1}$}
                                 node[below, font=\scriptsize, text=red!60!black]{shift $+1$, no stop-grad} (tproj.south);

    \draw[densely dashed, line width=0.9pt, black!55]
         (pproj.east) -- node[rotate=-90, font=\scriptsize, anchor=south]{shared weights} (tproj.west);

    \node[note] (lpred) at ($(pproj.north)!0.5!(tproj.north)+(1,0)$) {$\mathcal{L}_{\mathrm{pred}}$\\next-latent\\MSE};
    \draw[arrow] (pproj.north) -| (lpred.north);
    \draw[arrow] (tproj.north) -| (lpred.south);

    \node[note] (sig) at ($(pproj.north)!0.5!(tproj.north)+(2.5,0)$)
        {$\mathcal{L}_{\mathrm{SIGReg}}^{\mathrm{time}}$\\temporal isotropy\\regularization};
    \draw[arrow] (pproj.north) -| (sig.north);
    \draw[arrow] (tproj.north) -| (sig.south);

    \node[io, text width=3.6cm] (rep) at ($(vit.north)+(0,1.5)$)
        {\textbf{Representations}};
    \draw[arrow] (vit.north) -- (rep);

    \begin{scope}[on background layer]
      \node[draw=green!55!black, rounded corners, densely dashed, line width=1pt, inner sep=9pt,
            fit=(emb)(vit)] (encbox) {};
      \node[note, text=green!45!black, rotate=90, anchor=south] at ($(encbox.west)+(-0.05,0)$)
            {Encoder: use frozen after training for eval};
      \node[draw=orange!75!black, rounded corners, densely dashed, line width=1pt, inner sep=9pt,
            fit=(pproj)(tproj)(lpred)(sig)] (headbox) {};
      \node[note, text=orange!70!black, rotate=-90, anchor=south] at ($(headbox.east)+(0.05,0)$)
            {Training-only head: discarded after training};
    \end{scope}
\end{tikzpicture}%
}
\caption{LeNEPA training and what survives to evaluation. A causal ViT autoregressively predicts the next patch embedding: the top-layer prediction $\hat z_t$ and the shifted target $z_{t+1}$ are mapped by a \emph{shared} projector $h_\psi$ into the space where the next-latent MSE loss $\mathcal{L}_{\mathrm{pred}}$ is computed, with no stop-gradient on the target (stabilization comes from SIGReg, not EMA/stop-gradient). Temporal SIGReg $\mathcal{L}_{\mathrm{SIG}}^{\mathrm{time}}$ regularizes the projected tokens toward per-sample isotropy. The projector and losses (orange) are training-only and \emph{discarded} at evaluation; only the encoder (green) is \emph{kept}, and frozen probes read its best intermediate-layer representations.}
\Description{Block diagram of LeNEPA training. A time-series input passes through a convolutional patch embedding and a causal ViT. The top-layer prediction and the one-step-shifted target embedding are each passed through a shared projector, and a next-latent mean-squared-error loss compares them with no stop-gradient. Temporal SIGReg regularizes the projected tokens toward per-sample isotropy. A green dashed region marks the encoder, which is frozen and kept at evaluation and whose intermediate-layer representations are read by frozen probes; an orange dashed region marks the projector and losses, which are training-only and discarded at evaluation.}
\label{fig:lenepa_tower}
\end{figure*}
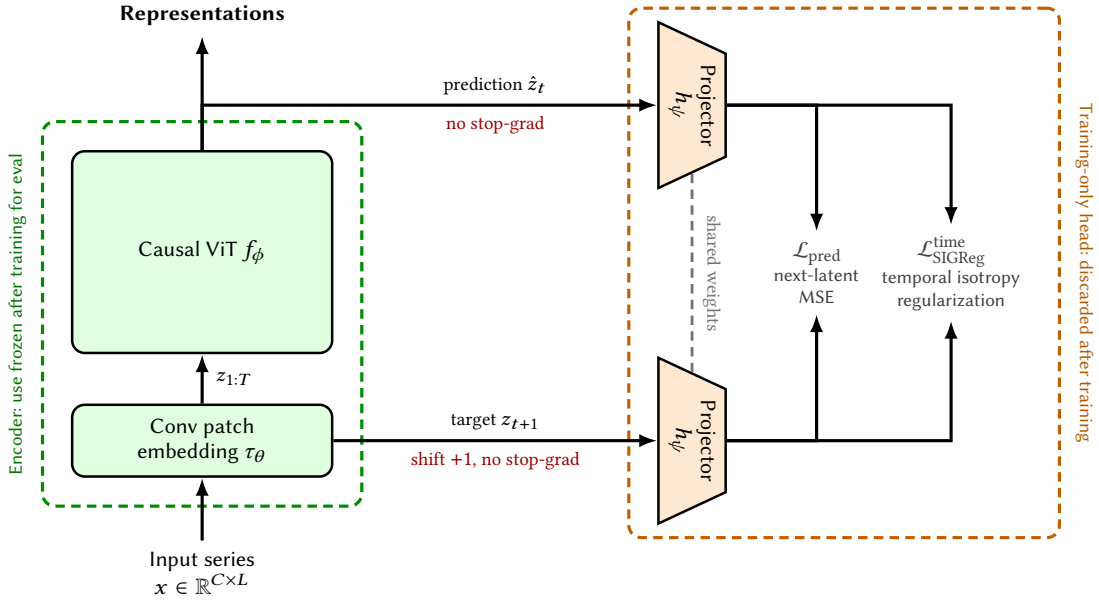

\section{LeNEPA -- lean no-augmentation SSL architecture}
\label{sec:method_lenepa}

We introduce \textbf{Latent Euclidean Next-Embedding Prediction Architecture (LeNEPA)}, a lean no-augmentation Self-Supervised Learning (SSL) architecture based on Next-Embedding Prediction Architecture (NEPA)~\citep{xu2025nextembeddingprediction}.
We split the input data (univariate or multivariate time series) into a sequence of ViT-style patch embeddings with strided convolution and process the sequence with a causal transformer backbone~\citep{vaswani2017attention}.
LeNEPA learns by autoregressively predicting the next latent token: at each index $t$, it predicts a token and minimizes the mean squared error to the target at $t{+}1$.

Vanilla NEPA stabilizes training with a teacher network updated by an exponential moving average (EMA) and by blocking gradients through the targets; LeNEPA replaces this mechanism with \textbf{Sketched Isotropic Gaussian Regularization (SIGReg)}~\citep{balestriero2025lejepa}, which regularizes the token embedding distribution toward an isotropic Gaussian.
Following Guillotine Regularization~\citep{bordes2022guillotine}, we compute the prediction loss in a projected space and discard the projector at evaluation time.
Let $x\in\mathbb{R}^{B\times C\times L}$, let $z=\tau_{\theta}(x)\in\mathbb{R}^{B\times T\times D}$ be strided-convolution patch embeddings, and let $\hat z=f_{\phi}(z)$ be causal transformer outputs, where $\hat z_{b,t}$ depends only on $z_{b,1:t}$.
With projector $h_{\psi}:\mathbb{R}^{D}\rightarrow\mathbb{R}^{d}$, the LeNEPA prediction loss is
\begin{equation}
  \mathcal{L}_{\mathrm{pred}}
  =
  \frac{1}{B(T-1)}\sum_{b=1}^{B}\sum_{t=1}^{T-1}
  \left\|h_{\psi}(\hat z_{b,t}) - h_{\psi}(z_{b,t+1})\right\|_2^2 .
\label{eq:main_lenepa_pred}
\end{equation}
Unlike NEPA, LeNEPA does not stop gradients through the target tokens in Eq.~\eqref{eq:main_lenepa_pred}; stabilization comes from SIGReg.

For time series, global non-collapse is insufficient because tokens can also collapse along the temporal axis within each sample.
Batch-wise or pooled SIGReg can keep an aggregate representation non-degenerate while still allowing collapse over the temporal axis; temporal SIGReg instead regularizes the token set inside each sample, directly counterbalancing the next-token regression pressure.
Let $u^{(\ell)}\in\mathbb{R}^{B\times T\times d}$ denote projected patch-token embeddings from layer $\ell$.
Unlike SIGReg regularization in the original LeJEPA design, we find that SIGReg over the temporal axis works best as a LeNEPA stabilizer: it regularizes the set of patch tokens within each sample,
\begin{equation}
  \mathcal{L}_{\mathrm{SIG}}^{\mathrm{time}}
  =
  \frac{1}{B|\mathcal{L}_{\mathrm{T}}|}\sum_{\ell\in\mathcal{L}_{\mathrm{T}}}\sum_{b=1}^{B}
  \mathrm{SIGReg}\!\left(\{u^{(\ell)}_{b,t}\}_{t=1}^{T}\right).
\label{eq:main_sigreg_time}
\end{equation}
The main objective is
\begin{equation}
  \mathcal{L}
  =
  \lambda_{\mathrm{pred}}\mathcal{L}_{\mathrm{pred}}
  +
  \lambda_{\mathrm{T}}\mathcal{L}_{\mathrm{SIG}}^{\mathrm{time}},
\label{eq:main_lenepa_objective}
\end{equation}
with $\lambda_{\mathrm{T}}=20$ and $\mathcal{L}_{\mathrm{T}}=\{0,8\}$ in the main PTB-XL/Diag configuration;
this setting was selected from ablations run on PTB-XL (Table~\ref{tab:ablation_summary}).
We also evaluated batch-wise (B), pooled-representation (R), innovation/difference (I), and combined BRI placements; temporal (T) SIGReg was the only single-component placement with sustained frozen-backbone gains across both datasets.

\begin{lstlisting}[language=Python,caption={Compact LeNEPA training step},label={lst:main_lenepa_train}]
# x: [B, C, L]
z = patch_embed(x)                         # [B, T, D]
tokens = causal_transformer(z)              # list of per-layer [B, T, D]
z_hat = tokens[-1]                          # top-layer predictions

pred = projector(z_hat[:, :-1])
target = projector(z[:, 1:])                # no stop-gradient in LeNEPA
loss = lambda_pred * mse(pred, target)

u_T = [projector(tokens[l]) for l in sigreg_time_layers]
loss += lambda_T * SIGReg_time_per_sample(u_T)

loss.backward()
optimizer.step()
\end{lstlisting}

\begin{figure*}[p]
\centering
\includegraphics[width=\textwidth]{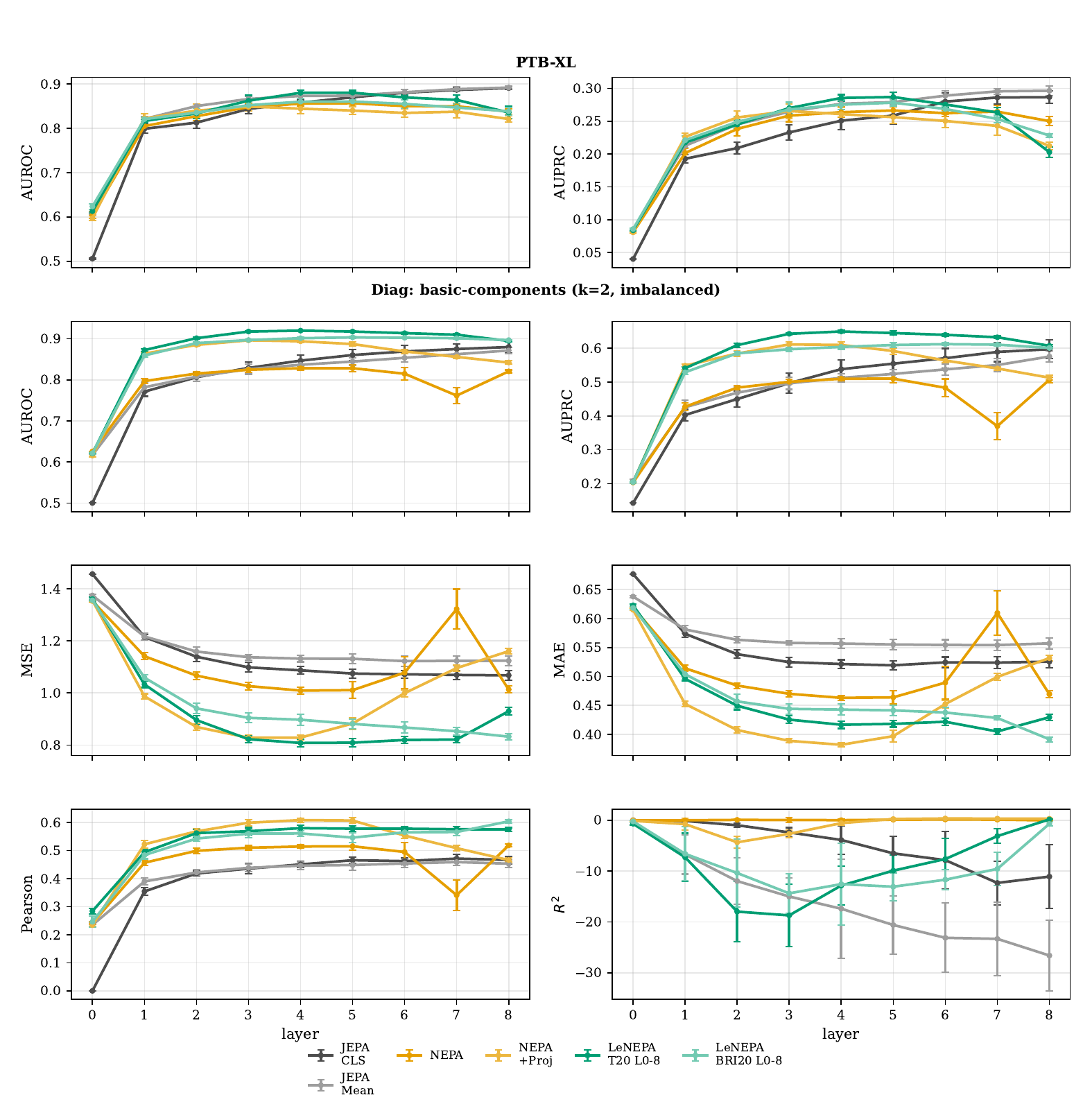}
\caption{Last-step (step $20{,}000$) layer-wise frozen-backbone probe results. We report PTB-XL (top row; AUROC/AUPRC) and Diag \emph{basic-components} results (bottom rows; AUROC/AUPRC and MSE/MAE/Pearson/$R^2$). Each line corresponds to a method; points denote probed layers $0$--$8$.}
\Description{Line plots of frozen-backbone probe performance across backbone layers for PTB-XL and Diag diagnostic metrics.}
\label{fig:last_step_by_layer_lines}
\end{figure*}

When evaluating frozen backbones, we find that intermediate layers provide stronger representations than the final layer for NEPA and LeNEPA (Section~\ref{sec:ablations_diagnostics}; Figure~\ref{fig:last_step_by_layer_lines}).
This is expected for next-embedding objectives: upper backbone blocks can partially act as an implicit predictor that maps features into the next-token regression space.
Therefore, we probe all backbone layers; headline LeNEPA classification readouts use a fixed mid-layer, while best-layer values are retained for baselines and diagnostic analyses where they should be read as oracle upper bounds.

\section{Evaluation Protocol}
\label{sec:evaluation_protocol}

We evaluate representations with frozen-backbone probes because our goal is to test whether a pretraining recipe produces useful sequence embeddings when reused unchanged.
Throughout the paper, \textbf{LeNEPA} denotes the objective and architecture, not a single shared checkpoint or foundation encoder.
We use three separately pretrained LeNEPA encoders: \textbf{LeNEPA-PTBXL}, trained on PTB-XL~\citep{wagner2020ptbxl}; \textbf{LeNEPA-Diag}, trained on Diag, a synthetic time-series corpus generated with Aionoscope~\citep{chemeris2026aionoscope}; and \textbf{LeNEPA-CauKer}, trained on CauKer~\citep{xie2025cauker}.
These checkpoints share the same no-augmentation latent-prediction recipe, but their weights are trained independently.

We separate two evaluation questions to avoid conflating recipe reuse with checkpoint transfer.
\textbf{Fixed-recipe pretraining reuse} asks whether the same SSL objective configuration can be reused when the pretraining distribution changes, without re-designing method-specific augmentations or loss hyperparameters.
We test this by training both methods independently on both datasets: LeNEPA-PTBXL and LeNEPA-Diag use exactly the same no-augmentation objective configuration, while JEPA-PTBXL and JEPA-Diag use the same ECG-tuned view recipe. In both cases, only the pretraining dataset changes.
Crucially, both recipes were developed and fixed using PTB-XL experiments: the JEPA masking schedule was tuned for ECG morphology, and the LeNEPA SIGReg configuration ($\lambda_{\mathrm{T}}=20$, layers $\{0,8\}$) was likewise selected from PTB-XL ablations.
Neither recipe was optimized for Diag.
We hypothesize that masking recipes encode domain-semantic assumptions---which temporal regions constitute meaningful prediction targets---whereas regularization hyperparameters primarily encode optimization dynamics; moving from a masking-based view recipe to an augmentation-free regularization-based recipe should therefore reduce the domain-specific tuning required when the signal family changes.
This protocol does not test the best possible Diag-tuned JEPA configuration; a JEPA recipe designed for Diag's signal characteristics would likely recover strong performance. Rather, the experiment measures the cost of relying on ECG-masking assumptions when that recipe is reused on a structurally different signal family.
\textbf{Frozen-encoder checking} asks whether one pretrained encoder can produce useful features on new downstream datasets without changing its weights.
We test this by training LeNEPA-CauKer once, freezing it, extracting per-layer embeddings, and evaluating Random-Forest probes on UCR-128, following the MantisV2 testing protocol~\citep{feofanov2026mantisv2}.

\begin{table}[t]
\centering
\caption{Pretrained encoder instances used in the evaluation. Each row is a separately pretrained checkpoint.}
\label{tab:lenepa_instances}
\small
\setlength{\tabcolsep}{3pt}
\begin{tabularx}{\linewidth}{l X X X}
\toprule
Encoder & Pretraining & Evaluation & Purpose \\
\midrule
LeNEPA-PTBXL
  & PTB-XL
  & PTB-XL
  & In-domain ECG reference against ECG-tuned JEPA \\
LeNEPA-Diag
  & Diag
  & Diag
  & Fixed-recipe reuse across signal families \\
JEPA-PTBXL
  & PTB-XL
  & PTB-XL
  & ECG-tuned JEPA reference under the same fixed-horizon protocol \\
JEPA-Diag
  & Diag
  & Diag
  & Same ECG-tuned JEPA recipe retrained on Diag to test unchanged reuse \\
LeNEPA-CauKer
  & CauKer
  & UCR-128
  & Frozen-encoder check on external classification datasets \\
\bottomrule
\end{tabularx}
\end{table}

Table~\ref{tab:reproducibility_config} summarizes the configuration used for the headline experiments.
Table~\ref{tab:reproducibility_config} and Listing~\ref{lst:main_lenepa_train} provide the information needed to reproduce the headline experiments.

\begin{table*}[t]
\centering
\scriptsize
\caption{Implementation and evaluation configuration for the main experiments. PTB-XL and Diag use the same fixed-horizon recipe-reuse matrix; the CauKer/UCR row documents the frozen-encoder check in Table~\ref{tab:ucr_mantis_comparison}.}
\label{tab:reproducibility_config}
\setlength{\tabcolsep}{3pt}
\begin{tabularx}{\textwidth}{p{0.16\textwidth} X X}
\toprule
Item & PTB-XL / Diag fixed-horizon matrix & CauKer $\rightarrow$ UCR-128 frozen-encoder check \\
\midrule
Training horizon and seeds
& $20{,}000$ gradient steps; seeds $s\in\{0,1,2,3,4\}$; offline probes every $1{,}000$ steps, including step $0$
& $20{,}000$ gradient steps; seed $0$; UCR evaluated at each saved checkpoint and reported at the best and final checkpoints \\
Optimizer
& AdamW, batch size $256$, bfloat16; cosine learning-rate schedule $10^{-4}\rightarrow10^{-6}$ with $1{,}000$ warmup steps; weight decay $0.01\rightarrow0.1$; betas $(0.9,0.99)$
& AdamW, batch size $256$, bfloat16; cosine learning-rate schedule $2{\times}10^{-4}\rightarrow2{\times}10^{-4}$ with $1{,}000$ warmup steps; weight decay $0.01\rightarrow0.1$; betas $(0.9,0.99)$ \\
Backbone and tokenizer
& Causal ViT-XS, depth $8$, $4$ heads, MLP ratio $4$, QK-Norm $(\epsilon=10^{-6})$, SwiGLU, RoPE; Conv1d patch embedding with $P=25$ for length-$5000$ signals; $D=192$; mean-pooling readout
& Causal ViT-XS, depth $8$, $4$ heads, MLP ratio $4$, QK-Norm, SwiGLU, RoPE; length-$5000$ CauKer sequences; $D=256$; sequence-normalized scalar-statistics path added for the Mantis-facing UCR comparison \\
LeNEPA loss
& Projected MSE next-token loss, no stop-gradient, $\lambda_{\mathrm{pred}}=1$; temporal SIGReg with $\lambda_{\mathrm{T}}=20$ on layers $\{0,8\}$; projector is MLP+BN+ReLU with output dimension $64$
& Same projected next-token objective; temporal SIGReg on layers $\{0,8\}$ with $\lambda_{\mathrm{T}}=2.5$; projector output dimension $64$ \\
JEPA view recipe
& Fixed ECG-tuned masking view recipe with keep ratio $\rho\sim\mathrm{Uniform}(0.15,0.25)$; JEPA evaluated with both CLS-token and mean-pooling readouts
& Not used \\
Probe protocol
& Frozen per-layer linear probes on layers $0$--$8$; $5{,}000$ probe steps; AdamW, learning rate $0.01$, weight decay $0$, batch size $256$; metrics aggregated as median and sample standard deviation across seeds
& Frozen per-layer embeddings on UCR-128; Random Forest classifier with $200$ trees and all 128 train/test splits; mean accuracy averaged across datasets \\
\bottomrule
\end{tabularx}
\end{table*}

\begin{figure*}[t]
\centering
\includegraphics[width=\textwidth]{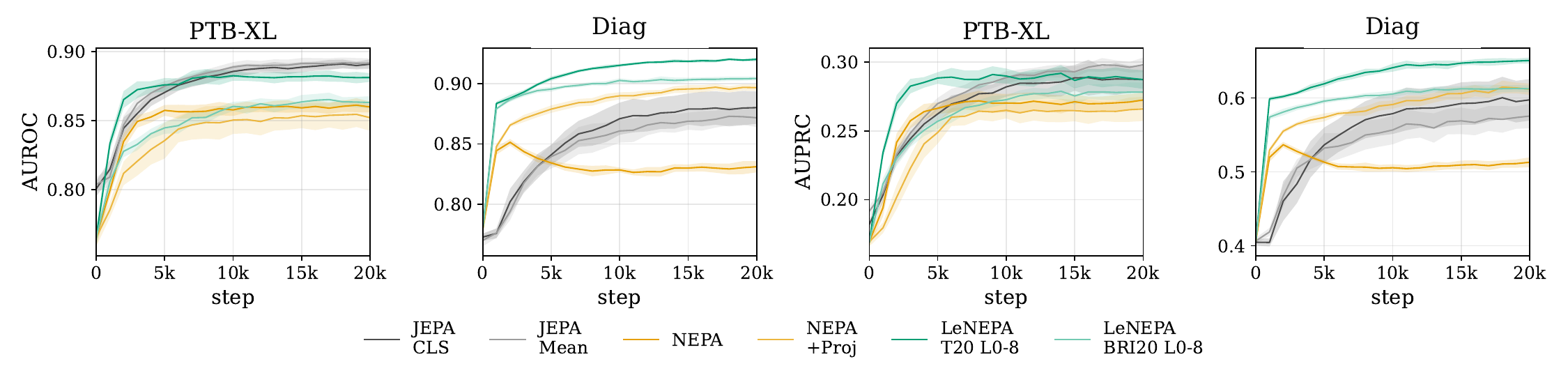}
\caption{PTB-XL/Diag fixed-recipe reuse experiment: best-layer frozen-backbone classification dynamics over the fixed $20{,}000$-step pretraining horizon. Curves show AUROC/AUPRC at each offline-probe evaluation step for PTB-XL and Diag. This tests recipe portability, not shared-checkpoint transfer or a claim that Diag-tuned JEPA cannot work. Both methods are trained separately on each dataset; the fixed ECG-tuned JEPA recipe remains strong in-domain but is weaker when reused unchanged on Diag.}
\Description{Four line plots showing AUROC and AUPRC over pretraining steps for PTB-XL and Diag. LeNEPA improves rapidly and remains strong on Diag, while the ECG-tuned JEPA recipe has weaker gains when reused unchanged on Diag.}
\label{fig:dynamics_best_layer_classification_transfer}
\end{figure*}

For the NEPA/LeNEPA/JEPA comparisons on PTB-XL and Diag, all methods use the same ViT-XS backbone size and are trained for a fixed horizon of $20{,}000$ gradient steps.
For NEPA/LeNEPA, we use mean pooling over causal patch-token representations; for JEPA, we report both CLS-token and mean-pooling readouts to control for readout mismatch.
Because next-embedding objectives can place the most probe-useful representation at an intermediate depth, we train offline linear probes on frozen per-layer representations for layers $0$--$8$.
For the last-step headline LeNEPA classification readouts, we use a fixed mid-layer $L4$ representation rather than an oracle layer; this is conservative on PTB-XL and exactly matches the best layer on Diag and the best UCR checkpoint (Table~\ref{tab:layer_selection_sensitivity}).
For JEPA/NEPA baselines and dense Diag diagnostics, we retain best-layer values to avoid penalizing baselines for their different depth profiles and to keep dense probes as diagnostic rather than headline comparison metrics.
We evaluate probes every $1{,}000$ pretraining steps, including step $0$, and report last-step values unless stated otherwise.
For PTB-XL and Diag, each objective--dataset pair is repeated with five random seeds; tables and plots in this paper aggregate per-seed metrics as median $\pm$ sample standard deviation.
For delta summaries, we compute each seed's difference between step $20{,}000$ and step $0$, selecting the best layer independently at each step.
For early-gain summaries on AUROC/AUPRC, define the normalized gain
\begin{equation}
  r_m(t)=\frac{m(t)-m(0)}{m(20{,}000)-m(0)}
\end{equation}
for metric $m$ at pretraining step $t$, and let $\tau_q(m)$ be the first offline-probe evaluation step at which $r_m(t)\ge q$.
We compute these thresholds on the median learning curves in Figure~\ref{fig:dynamics_best_layer_classification_transfer}; because probes are evaluated every $1{,}000$ steps, $\tau_q$ is a discrete step count rather than an interpolated time.

PTB-XL is a 12-lead ECG classification dataset with clinically meaningful multi-label targets~\citep{wagner2020ptbxl}.
This benchmark is a useful in-domain test because ECG\_JEPA is a strong baseline designed around ECG-specific JEPA augmentations~\citep{weimann2024jepaecg}.
To test fixed-recipe reuse across a different signal family, we construct \textbf{Diag}, a synthetic diagnostic corpus generated with the Aionoscope library~\citep{chemeris2026aionoscope}, implemented as an online generator rather than a fixed finite dataset.
Aionoscope provides a Process-to-View generator that renders latent process states into observed time series and emits exact categorical and dense labels from the same state.
Our Diag stream uses Aionoscope's Primitive Process Mixtures generator, but it is a LeNEPA evaluation stream rather than the public Aionoscope benchmark sweep: here we use length-$5000$ univariate sequences, exactly $2$ active components per sample, and a custom imbalanced component sampler.
Each sample is generated by selecting active latent components from a fixed inventory of $14$ component types spanning constant baselines, noise processes, trends, periodic waveforms, and sparse events.
The main stream is intentionally imbalanced: most components have sampling weight $1.0$, while six rare components have weight $0.02$ (\texttt{gaussian\_noise}, \texttt{log\_trend}, \texttt{square}, \texttt{spike}, \texttt{level\_change}, and \texttt{gaussian}).

Diag exposes two kinds of targets used for frozen-backbone probing.
Categorical targets are multi-hot indicators of the active component types and are evaluated with macro AUROC/AUPRC.
Dense targets are scalar generative parameters recorded by the generator metadata, including noise scale, trend coefficients, periodic amplitude/frequency/phase/offset, and event time/amplitude/width; disabled components are masked for the corresponding dense target.
SSL training never uses these targets.
Pretraining and probing streams use separate fixed RNG streams, so probes evaluate fresh generated samples from the same distribution rather than memorized training examples.

Finally, we include a frozen-encoder check by testing LeNEPA-CauKer trained on CauKer~\citep{xie2025cauker} on the full UCR archive of 128 univariate classification datasets~\citep{dau2019ucr}.
Unlike the PTB-XL/Diag fixed-recipe study, the CauKer-to-UCR experiment evaluates a single frozen LeNEPA checkpoint across 128 external datasets.
For this experiment, we follow the Mantis frozen-feature protocol: extract frozen per-layer sequence embeddings, train a Random Forest classifier on each dataset's training split, evaluate on its test split, and average accuracy across datasets~\citep{feofanov2026mantisv2,xie2025cauker}.
We use the original Mantis Random-Forest result as the primary published anchor, and report NuTime and MOMENT only as protocol-matched context from the MantisV2 Random-Forest comparison~\citep{lin2024nutime,goswami2024moment}.

\begin{table*}[t]
\centering
\small
\caption{PTB-XL/Diag fixed-horizon experiment matrix at the last step ($20{,}000$ updates) for the controlled five-seed comparisons. JEPA/NEPA rows report best-layer frozen-backbone probes; LeNEPA classification columns use a fixed $L4$ readout, while LeNEPA dense Diag columns report metric-specific best-layer diagnostic probes. Values are aggregated across seeds as $\mathrm{median}(\mathrm{std})$, with std shown in $10^{-3}$ units.}
\label{tab:summary_selected_best_last_step}
\begin{tabular*}{\textwidth}{@{\extracolsep{\fill}}llcccccccc@{}}
\toprule
\multicolumn{2}{c}{} & \multicolumn{2}{c}{PTB-XL} & \multicolumn{6}{c}{Diag} \\
\cmidrule(lr){3-4} \cmidrule(lr){5-10}
model & config & \multicolumn{1}{c}{AUROC $\uparrow$} & \multicolumn{1}{c}{AUPRC $\uparrow$} & \multicolumn{1}{c}{AUROC $\uparrow$} & \multicolumn{1}{c}{AUPRC $\uparrow$} & \multicolumn{1}{c}{MSE $\downarrow$} & \multicolumn{1}{c}{MAE $\downarrow$} & \multicolumn{1}{c}{Pearson $\uparrow$} & \multicolumn{1}{c}{$R^2$ $\uparrow$} \\
\midrule
JEPA & CLS & \textbf{0.891(3)} & \underline{0.287(9)} & 0.880(13) & 0.597(28) & 1.065(13) & 0.515(5) & 0.480(12) & -0.005(24) \\
JEPA & Mean & \textbf{0.892(3)} & \textbf{0.298(5)} & 0.872(7) & 0.576(16) & 1.113(13) & 0.548(5) & 0.462(11) & -0.514(291) \\
NEPA &  & 0.860(3) & 0.272(5) & 0.831(4) & 0.513(6) & 1.001(9) & 0.461(3) & 0.522(4) & \underline{0.197(24)} \\
NEPA & +Proj & 0.852(9) & 0.266(8) & \underline{0.897(2)} & \underline{0.612(7)} & \underline{0.827(8)} & \textbf{0.382(3)} & \textbf{0.610(7)} & \textbf{0.348(23)} \\
LeNEPA & T20 L0-8 & \underline{0.880(6)} & \underline{0.285(5)} & \textbf{0.920(1)} & \textbf{0.650(4)} & \textbf{0.797(12)} & \underline{0.405(4)} & \underline{0.584(6)} & 0.164(105) \\
\bottomrule
\end{tabular*}
\end{table*}

\begin{table*}[t]
\centering
\caption{UCR-128 frozen-encoder check. We report mean accuracy over all 128 UCR datasets using frozen sequence embeddings and a Random Forest classifier. The table lists each model's pretraining data because baselines differ in architecture, objective, and whether UCR/UEA training data or larger heterogeneous corpora were used. The LeNEPA-CauKer row reports the best checkpoint over the fixed $20{,}000$-step trajectory; Mantis is the closest published CauKer-only Mantis-family anchor, and NuTime/MOMENT provide protocol-matched context from the same Random-Forest comparison.}
\label{tab:ucr_mantis_comparison}
\small
\setlength{\tabcolsep}{3pt}
\begin{tabularx}{\textwidth}{l X l l X X r}
\toprule
Model & Pretrain data & Params & Attention & Tokenizer / input engineering & SSL objective & Acc. (\%) \\
\midrule
LeNEPA
  & CauKer; no UCR
  & $6.3$M
  & causal
  & Conv patches + global stats
  & latent prediction; no aug.
  & $\mathbf{77.65}$ \\
Mantis
  & CauKer; no UCR
  & $8.1$M
  & bidirectional
  & Conv patches + patch stats + diff
  & contrastive; crop aug.
  & $78.81$ \\
NuTime~\citep{lin2024nutime}
  & incl. UCR/UEA train
  & $2.0$M
  & bidirectional
  & Window shape + window stats
  & BYOL; crop aug.
  & $77.32$ \\
MOMENT~\citep{goswami2024moment}
  & incl. UCR/UEA train
  & $161$M
  & bidirectional
  & T5 patches
  & masked reconstruction
  & $77.89$ \\
\bottomrule
\end{tabularx}
\vspace{0.25em}
\footnotesize
LeNEPA-CauKer peaks at $77.65\%$ at $17{,}000$ steps and remains at $77.20\%$ at the final $20{,}000$-step checkpoint (Figure~\ref{fig:ucr_layer_profile_lenepa_seqnorm}).
LeNEPA-CauKer row adds MSSE for this Mantis-facing UCR comparison because Mantis itself uses an MSSE-encoded scalar-statistics path; the main LeNEPA experiments elsewhere in the paper do not rely on MSSE.
NuTime/MOMENT entries use the Random-Forest UCR-128 averages reported in MantisV2~\citep{feofanov2026mantisv2}.
\end{table*}

\section{Experiments}
\label{sec:experiments}

\begin{figure}[t]
\centering
\includegraphics[width=\linewidth]{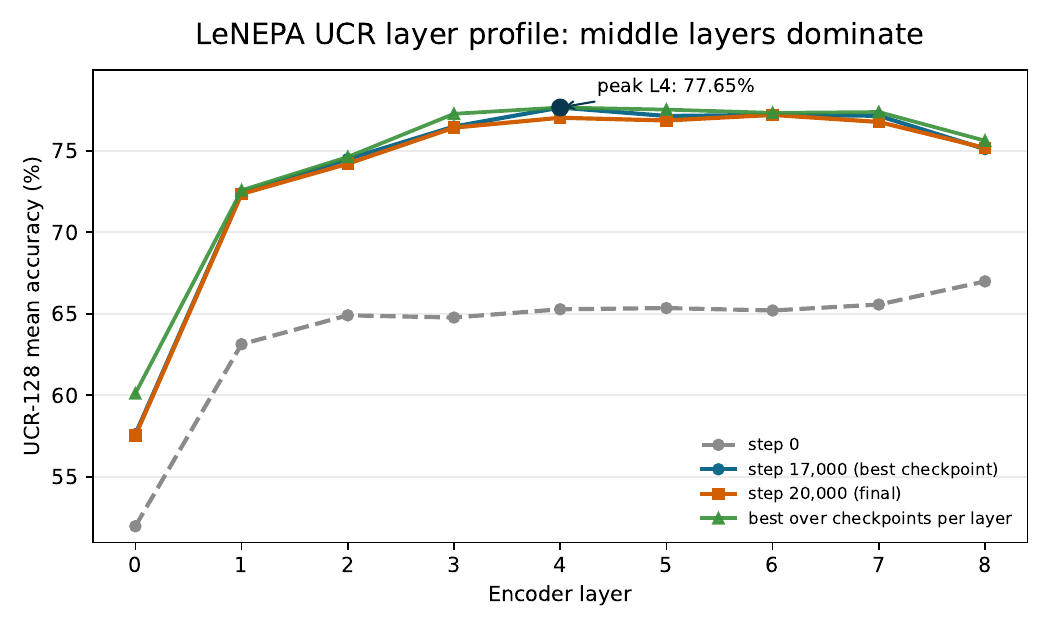}
\caption{UCR-128 layer-profile diagnostic for the LeNEPA-CauKer run used in Table~\ref{tab:ucr_mantis_comparison}. Each point is mean Random-Forest accuracy over the 128 UCR datasets for one frozen backbone layer. Intermediate layers outperform the final layer: the run peaks at $77.65\%$ at $17{,}000$ steps ($L4$) and remains at $77.20\%$ at the final $20{,}000$-step checkpoint ($L6$).}
\Description{Line plot of UCR-128 Random Forest accuracy by LeNEPA layer for selected checkpoints. Intermediate layers peak above the final layer.}
\label{fig:ucr_layer_profile_lenepa_seqnorm}
\end{figure}

We evaluate frozen-backbone representations using per-layer probes.
For time series, we evaluate fixed-recipe reuse on PTB-XL (12-lead ECG classification) and Diag (classification + dense regression), then use UCR-128 as a separate frozen-encoder check.

\paragraph{Fixed-recipe reuse: PTB-XL vs. Diag.}
The baseline JEPA model~\citep{weimann2024jepaecg} is tuned for ECG and relies on dataset-specific augmentations, and thus performs well on the dataset it was designed for.
In this in-domain regime, LeNEPA remains competitive, showing that eliminating augmentations does not preclude useful frozen features.
We then test fixed-recipe reuse by training both LeNEPA and JEPA again on Diag while keeping each method's recipe unchanged.
As shown by Figure~\ref{fig:dynamics_best_layer_classification_transfer}, the JEPA configuration that is strong on PTB-XL is weaker when reused unchanged on Diag: JEPA-Diag's training gains shrink markedly, whereas LeNEPA-Diag retains strong training gains and beats JEPA-Diag both in final performance and training dynamics.
This does not rule out a Diag-tuned JEPA recipe; a JEPA recipe designed for Diag's signal characteristics would likely recover competitive performance. Rather, the experiment identifies \emph{ECG-masking assumptions}---the semantic tie between JEPA's view recipe and ECG morphology---as a source of fragility when that recipe is reused on a structurally different signal family.
A general-purpose JEPA masking recipe might transfer better, but would still require per-domain view engineering: exactly the cost that a no-augmentation objective avoids by design.
Finally, Figure~\ref{fig:dynamics_best_layer_classification_transfer} summarizes early learning dynamics under the time-to-gain definition in Section~\ref{sec:evaluation_protocol}.
On the median AUROC/AUPRC curves, LeNEPA reaches $80\%$ of its final gain after $2$--$5$k updates across PTB-XL and Diag, whereas the faster JEPA readout reaches the same threshold after $5$--$10$k updates; because probes are evaluated every $1{,}000$ steps these thresholds are coarse discrete counts and per-seed timing varies.
The $90\%$ threshold is less uniform---LeNEPA reaches it after $5$--$10$k updates, while JEPA typically reaches it after $10$--$15$k updates---so we interpret these observations as qualitative evidence of faster early representation acquisition in this fixed-recipe protocol.

\paragraph{Frozen-encoder check: CauKer to UCR-128.}
We use UCR-128 as a frozen-encoder check that complements the fixed-recipe reuse study. We choose Mantis as the primary comparator because it is a state-of-the-art SSL/foundation architecture for time-series classification and is close to LeNEPA in the relevant evaluation regime: a frozen time-series encoder followed by a lightweight classifier~\citep{feofanov2025mantis,feofanov2026mantisv2}.
UCR-128 classification accuracy is Mantis's headline contribution; its architecture was purpose-built for this setting: bidirectional attention (full sequence context), a handcrafted first-difference branch, and a contrastive objective with Random Crop Resize augmentations.
The closest published anchor from this architecture uses frozen Mantis representations and reports $78.81\%$ average accuracy on all 128 UCR datasets with a Random Forest classifier.
As shown in Table~\ref{tab:ucr_mantis_comparison}, our LeNEPA-CauKer run peaks at $77.65\%$ over the fixed $20$k-step pretraining trajectory (single seed) and remains at $77.20\%$ at the final checkpoint (Figure~\ref{fig:ucr_layer_profile_lenepa_seqnorm}).
We treat this single-seed result as an existence proof: a LeNEPA encoder with causal attention, no augmentations, and no first-difference branch---each a structural disadvantage relative to Mantis for UCR classification---can already produce frozen features competitive with a model that was specifically designed and optimised for this benchmark.
Additional seeds would characterise variance; this run establishes that the no-augmentation causal recipe can reach the Mantis neighbourhood without any UCR-specific engineering.
NuTime and MOMENT in the same MantisV2 Random-Forest UCR table are reported as $77.32\%$ and $77.89\%$, respectively; the best LeNEPA checkpoint is above NuTime and within $0.24$ percentage points of MOMENT~\citep{lin2024nutime,goswami2024moment}.
Because this comparison uses a single LeNEPA pretraining seed, a best checkpoint over the trajectory, and baselines with different pretraining corpora and architectures, we treat it as protocol-matched context for the frozen features.
We therefore do not treat stronger MantisV2 final-comparison or leaderboard numbers as direct baselines, because they are designed for leaderboard performance through heavier model and test-time/evaluation choices rather than for isolating the effect of a no-augmentation objective.

The following diagnostics summarize the main ablations needed to interpret the results.

\section{Ablations and Diagnostics}
\label{sec:ablations_diagnostics}

Diag acts as diagnostic instrumentation rather than as a paper contribution.
Each generated sequence has known latent factors and categorical/dense targets; SSL pretraining does not use these labels, and they are used only after freezing the backbone to probe which factors are linearly recoverable from each layer.

Table~\ref{tab:ablation_summary} summarizes the controlled ablations behind the main LeNEPA design choices.
Because the page limit prevents full ablation tables, we report the design-level conclusions here.
The projector rows test the Guillotine Regularization idea that SSL losses can be applied in a disposable head while evaluating the truncated backbone~\citep{bordes2022guillotine}.
As Table~\ref{tab:summary_selected_best_last_step} shows, NEPA with a projector (NEPA\,+\,Proj) is already a strong baseline: it outperforms LeNEPA on several dense Diag metrics (MAE, Pearson, $R^2$) while LeNEPA leads on classification (AUROC, AUPRC) and MSE.
The contribution of temporal SIGReg over NEPA\,+\,Proj is therefore most clearly visible on classification and training dynamics, while the dense regression picture is more mixed.

\begin{table*}[t]
\centering
\small
\caption{Compact ablation summary supporting the LeNEPA design choices.}
\label{tab:ablation_summary}
\setlength{\tabcolsep}{4pt}
\renewcommand{\arraystretch}{1.12}
\begin{tabularx}{\textwidth}{p{0.18\textwidth} p{0.30\textwidth} p{0.25\textwidth} X}
\toprule
Design question & Controlled ablation & Main observation & Implication \\
\midrule
Projector loss space
& Guillotine-style projector: NEPA/LeNEPA trained with and without the projector
& Projector-enabled variants improve $22/24$ last-step probe comparisons.
& Compute prediction and SIGReg losses in projected space, then discard the projector for evaluation. \\
\addlinespace
Projector capacity
& Output width, hidden width, and MLP depth of the Guillotine-style projector
& Larger or deeper projectors do not improve performance over the fixed projector used in the main runs.
& The main projector effect comes from separating the loss space from the evaluated backbone space, not from adding projector capacity. \\
\addlinespace
SIGReg axis
& Temporal (T) SIGReg versus batch-wise (B), pooled-representation (R), innovation/difference (I), and batch-time (BT) placements
& Temporal SIGReg is the only single-component placement with sustained frozen-backbone gains across PTB-XL and Diag.
& Use Temporal SIGReg as the LeNEPA anti-collapse stabilizer. \\
\addlinespace
SIGReg layer placement
& Temporal SIGReg on both prediction/target sides $\{0,8\}$ vs.\ one side or inner layers
& Regularizing both sides of the prediction relation is more stable in our fixed-recipe runs; target-only or off-path regularization can degrade or collapse.
& Apply SIGReg where it directly counterbalances the next-embedding prediction loss. \\
\addlinespace
JEPA view recipe
& PTB-XL JEPA keep-ratio $(0.15,0.25)$ vs.\ $(0.05,0.10)$ and $(0.30,0.40)$
& A $2$--$3\times$ change in masking strength substantially degrades probe dynamics.
& Augmentation/view hyperparameters can be sensitive, motivating a no-augmentation objective. \\
\bottomrule
\end{tabularx}
\end{table*}

Finally, layer-wise probing reveals a pattern that is increasingly important for latent-prediction encoders.
Across both datasets, JEPA probe performance tends to improve with depth and peak near the top layers.
In contrast, NEPA and LeNEPA often achieve their best probe scores at intermediate layers (frequently around $l\approx3$--$5$), with the final layer sometimes underperforming these mid-level representations.
This suggests that the deepest blocks of next-embedding predictors can partially specialize to the prediction task, while intermediate layers retain more probe-useful features.

\paragraph{Layer-selection sensitivity.}
The best-layer protocol is an oracle upper bound.
The key question is therefore whether this oracle choice materially changes the conclusions.
We check this by comparing oracle selection to a fixed mid-layer readout.
Table~\ref{tab:layer_selection_sensitivity} compares the oracle best layer with a fixed $L4$ readout and the final transformer layer $L8$ for LeNEPA classification results.
The classification optima are concentrated in the middle of the backbone: PTB-XL peaks at $L5$, while Diag and the best UCR checkpoint peak at $L4$.
Across the five-seed PTB-XL and five-seed Diag classification evaluations, using the fixed $L4$ readout removes almost all of the oracle advantage, whereas using the final layer can be substantially worse.
Thus, the LeNEPA classification conclusions do not rely on per-dataset layer tuning; dense Diag probes are more metric-dependent and should be read primarily as diagnostics of where information appears in the backbone.

\begin{table}[t]
\centering
\scriptsize
\caption{Layer-selection sensitivity motivating the fixed-$L4$ LeNEPA classification readout. Best-layer values are oracle selections over probed layers; fixed $L4$ uses one mid-layer readout across datasets; $L8$ is the final transformer layer. Parentheses report the difference from the oracle value.}
\label{tab:layer_selection_sensitivity}
\begin{tabular}{llrrrr}
\toprule
Dataset & Metric & Oracle layer & Oracle & Fixed $L4$ ($\Delta$) & Final $L8$ ($\Delta$) \\
\midrule
PTB-XL & AUROC & $L5$ & $0.880$ & $0.880$ ($0.000$) & $0.835$ ($-0.045$) \\
PTB-XL & AUPRC & $L5$ & $0.287$ & $0.285$ ($-0.002$) & $0.203$ ($-0.084$) \\
Diag & AUROC & $L4$ & $0.920$ & $0.920$ ($0.000$) & $0.895$ ($-0.025$) \\
Diag & AUPRC & $L4$ & $0.650$ & $0.650$ ($0.000$) & $0.607$ ($-0.043$) \\
UCR-128 & Accuracy & $L4$ & $77.65$ & $77.65$ ($0.00$) & $75.10$ ($-2.55$) \\
\bottomrule
\end{tabular}
\end{table}

\section{Limitations and Future Work}
\label{sec:limitations}

While LeNEPA removes one source of time-series SSL engineering by avoiding augmentations, our current study tests recipe portability rather than claiming a foundation encoder.
The convolutional tokenizer still depends on dataset-specific design choices such as kernel size and stride, implicitly assumes regular sampling, and may limit transfer across signals with different temporal scales.
Although the PTB-XL/Diag fixed-recipe runs use the same temporal SIGReg setting, SIGReg still introduces scale and placement hyperparameters; the CauKer/UCR run uses a smaller temporal scale in a different tokenizer configuration, so principled defaults across dataset scales and tokenizers remain future work.
Diag is also a synthetic corpus generated with Aionoscope~\citep{chemeris2026aionoscope}, so conclusions drawn from it should be read as controlled state-accessibility evidence rather than as direct evidence of real-world diagnostic transfer.
Some diagnostic analyses and baseline comparisons still report best-layer values; those should be treated as oracle upper bounds unless a layer is fixed a priori or selected using labeled validation data.
For LeNEPA classification, the reported last-step readout is fixed to $L4$ and is stable relative to the oracle choice (Table~\ref{tab:layer_selection_sensitivity}).
The UCR-128 frozen-encoder check broadens the evidence beyond PTB-XL and Diag, but it is still univariate and uses one pretraining seed; broader UEA-style multivariate evaluations remain necessary.

\section{Conclusion}
\label{sec:conclusion}

Time-series SSL recipes are often entangled with view and augmentation choices that may be safe in one signal family but mismatched in another.
We studied this issue as a \emph{fixed-recipe stress test}: keep each method's SSL recipe unchanged, retrain on a new pretraining distribution, and evaluate whether frozen representations remain useful.
This setting is intentionally narrower than fully tuned benchmark comparison; it measures the cost of recipe reuse, not the best attainable performance after retuning.

We proposed \textbf{LeNEPA}, a no-augmentation next-embedding prediction architecture that replaces EMA/stop-gradient stabilization with temporal SIGReg and applies the predictive loss in a disposable projected space.
Under the PTB-XL/Diag fixed-horizon protocol, an ECG-tuned JEPA recipe remains strong in-domain but weakens when reused unchanged on Diag, while LeNEPA preserves useful frozen-probe gains under the same unchanged-recipe rule.
The learning curves also suggest faster early representation acquisition for LeNEPA in this protocol, though the evidence should be read as a fixed-recipe result rather than a claim about optimally tuned JEPA.

As a separate frozen-encoder check, a CauKer-pretrained LeNEPA checkpoint falls in the range of protocol-matched UCR-128 Random-Forest anchors while avoiding handcrafted augmentations and using a causal backbone.
This check provides external context for the learned features, but it is not a definitive foundation-model or leaderboard claim: it uses one pretraining seed, a best checkpoint over the trajectory, and a UCR-specific comparison setup.

We thus conclude that augmentation-free latent prediction is a promising building block for time-series representation learning when the operational constraint is to reuse an SSL recipe with minimal per-dataset view engineering.

\bibliographystyle{ACM-Reference-Format}
\bibliography{references}

@misc{weimann2024jepaecg,
title = {Self-Supervised Pre-Training with Joint-Embedding Predictive Architecture Boosts ECG Classification Performance},
author = {Weimann, Kuba and Conrad, Tim O. F.},
year = {2024},
note = {arXiv:2410.13867},
url = {https://arxiv.org/abs/2410.13867}
}

@misc{assran2023ijepa,
title = {Self-Supervised Learning from Images with a Joint-Embedding Predictive Architecture},
author = {Assran, Mahmoud and Duval, Quentin and Misra, Ishan and Bojanowski, Piotr and Vincent, Pascal and Rabbat, Michael and LeCun, Yann and Ballas, Nicolas},
year = {2023},
note = {arXiv:2301.08243},
url = {https://arxiv.org/abs/2301.08243}
}

@misc{xu2025nextembeddingprediction,
title = {Next-Embedding Prediction Makes Strong Vision Learners},
author = {Xu, Sihan and Ma, Ziqiao and Chai, Wenhao and Chen, Xuweiyi and Jin, Weiyang and Chai, Joyce and Xie, Saining and Yu, Stella X.},
year = {2025},
note = {arXiv:2512.16922},
url = {https://arxiv.org/abs/2512.16922}
}

@misc{balestriero2025lejepa,
title = {LeJEPA: Provable and Scalable Self-Supervised Learning Without the Heuristics},
author = {Balestriero, Randall and LeCun, Yann},
year = {2025},
note = {arXiv:2511.08544},
url = {https://arxiv.org/abs/2511.08544}
}

@misc{bordes2022guillotine,
title = {Guillotine Regularization: Why removing layers is needed to improve generalization in Self-Supervised Learning},
author = {Bordes, Florian and Balestriero, Randall and Garrido, Quentin and Bardes, Adrien and Vincent, Pascal},
year = {2022},
note = {arXiv:2206.13378},
url = {https://arxiv.org/abs/2206.13378}
}

@article{wagner2020ptbxl,
title = {PTB-XL, a large publicly available electrocardiography dataset},
author = {Wagner, Patrick and Strodthoff, Nils and Bousseljot, Ralf-Dieter and Kreiseler, Dieter and Lunze, Fatima I. and Samek, Wojciech and Schaeffter, Tobias},
journal = {Scientific Data},
volume = {7},
number = {1},
pages = {154},
year = {2020},
doi = {10.1038/s41597-020-0495-6},
url = {https://www.nature.com/articles/s41597-020-0495-6}
}

@misc{chen2020simclr,
title = {A Simple Framework for Contrastive Learning of Visual Representations},
author = {Ting Chen and Simon Kornblith and Mohammad Norouzi and Geoffrey Hinton},
year = {2020},
note = {arXiv:2002.05709},
url = {https://arxiv.org/abs/2002.05709}
}

@misc{he2020moco,
title = {Momentum Contrast for Unsupervised Visual Representation Learning},
author = {Kaiming He and Haoqi Fan and Yuxin Wu and Saining Xie and Ross Girshick},
year = {2020},
note = {arXiv:1911.05722},
url = {https://arxiv.org/abs/1911.05722}
}

@misc{he2021mae,
title = {Masked Autoencoders Are Scalable Vision Learners},
author = {Kaiming He and Xinlei Chen and Saining Xie and Yanghao Li and Piotr Doll{\'a}r and Ross Girshick},
year = {2021},
note = {arXiv:2111.06377},
url = {https://arxiv.org/abs/2111.06377}
}

@misc{grill2020byol,
title = {Bootstrap your own latent: A new approach to self-supervised Learning},
author = {Jean-Bastien Grill and Florian Strub and Florent Altch{\'e} and Corentin Tallec and Pierre H. Richemond and Elena Buchatskaya and Carl Doersch and Bernardo Avila Pires and Zhaohan Daniel Guo and Mohammad Gheshlaghi Azar and Bilal Piot and Koray Kavukcuoglu and R{\'e}mi Munos and Michal Valko},
year = {2020},
note = {arXiv:2006.07733},
url = {https://arxiv.org/abs/2006.07733}
}

@misc{caron2021dino,
title = {Emerging Properties in Self-Supervised Vision Transformers},
author = {Mathilde Caron and Hugo Touvron and Ishan Misra and Herv{\'e} J{\'e}gou and Julien Mairal and Piotr Bojanowski and Armand Joulin},
year = {2021},
note = {arXiv:2104.14294},
url = {https://arxiv.org/abs/2104.14294}
}

@misc{vaswani2017attention,
title = {Attention Is All You Need},
author = {Ashish Vaswani and Noam Shazeer and Niki Parmar and Jakob Uszkoreit and Llion Jones and Aidan N. Gomez and Lukasz Kaiser and Illia Polosukhin},
year = {2017},
note = {arXiv:1706.03762},
url = {https://arxiv.org/abs/1706.03762}
}

@misc{yue2022ts2vec,
title = {{TS2Vec}: Towards Universal Representation of Time Series},
author = {Zhihan Yue and Yujing Wang and Juanyong Duan and Tianmeng Yang and Congrui Huang and Yunhai Tong and Bixiong Xu},
year = {2022},
note = {arXiv:2106.10466},
url = {https://arxiv.org/abs/2106.10466}
}

@misc{eldele2021tstcc,
title = {Time-Series Representation Learning via Temporal and Contextual Contrasting},
author = {Emadeldeen Eldele and Mohamed Ragab and Zhenghua Chen and Min Wu and Chee Keong Kwoh and Xiaoli Li and Cuntai Guan},
year = {2021},
note = {arXiv:2106.14112},
url = {https://arxiv.org/abs/2106.14112}
}

@misc{woo2022cost,
title = {{CoST}: Contrastive Learning of Disentangled Seasonal-Trend Representations for Time Series Forecasting},
author = {Gerald Woo and Chenghao Liu and Doyen Sahoo and Akshat Kumar and Steven Hoi},
year = {2022},
note = {arXiv:2202.01575},
url = {https://arxiv.org/abs/2202.01575}
}

@misc{tonekaboni2021tnc,
title = {Unsupervised Representation Learning for Time Series with Temporal Neighborhood Coding},
author = {Sana Tonekaboni and Danny Eytan and Anna Goldenberg},
year = {2021},
note = {arXiv:2106.00750},
url = {https://arxiv.org/abs/2106.00750}
}

@misc{bardes2024vjepa,
title = {Revisiting Feature Prediction for Learning Visual Representations from Video},
author = {Adrien Bardes and Quentin Garrido and Jean Ponce and Xinlei Chen and Michael Rabbat and Yann LeCun and Mahmoud Assran and Nicolas Ballas},
year = {2024},
note = {arXiv:2404.08471},
url = {https://arxiv.org/abs/2404.08471}
}

@misc{liu2024tsaugselect,
title = {Guidelines for Augmentation Selection in Contrastive Learning for Time Series Classification},
author = {Liu, Ziyu and Alavi, Azadeh and Li, Minyi and Zhang, Xiang},
year = {2024},
note = {arXiv:2407.09336},
url = {https://arxiv.org/abs/2407.09336}
}

@article{dau2019ucr,
title = {The {UCR} time series archive},
author = {Dau, Hoang Anh and Bagnall, Anthony and Kamgar, Kaveh and Yeh, Chin-Chia Michael and Zhu, Yan and Gharghabi, Shaghayegh and Ratanamahatana, Chotirat Ann and Keogh, Eamonn},
journal = {IEEE/CAA Journal of Automatica Sinica},
volume = {6},
number = {6},
pages = {1293--1305},
year = {2019},
publisher = {IEEE}
}

@article{xie2025cauker,
title = {{CauKer}: classification time series foundation models can be pretrained on synthetic data only},
author = {Xie, Shifeng and Feofanov, Vasilii and Alonso, Marius and Odonnat, Ambroise and Zhang, Jianfeng and Palpanas, Themis and Redko, Ievgen},
journal = {arXiv preprint arXiv:2508.02879},
year = {2025}
}

@article{feofanov2025mantis,
title = {Mantis: Lightweight calibrated foundation model for user-friendly time series classification},
author = {Feofanov, Vasilii and Wen, Songkang and Alonso, Marius and Ilbert, Romain and Guo, Hongbo and Tiomoko, Malik and Pan, Lujia and Zhang, Jianfeng and Redko, Ievgen},
journal = {arXiv preprint arXiv:2502.15637},
year = {2025}
}

@misc{feofanov2026mantisv2,
title = {{MantisV2}: Closing the Zero-Shot Gap in Time Series Classification with Synthetic Data and Test-Time Strategies},
author = {Feofanov, Vasilii and Wen, Songkang and Zhang, Jianfeng and Pan, Lujia and Redko, Ievgen},
year = {2026},
note = {arXiv:2602.17868; ICLR 2026 TSALM Workshop Poster},
doi = {10.48550/arXiv.2602.17868},
url = {https://arxiv.org/abs/2602.17868}
}

@article{lin2024nutime,
title = {{NuTime}: Numerically Multi-Scaled Embedding for Large-Scale Time-Series Pretraining},
author = {Lin, Chenguo and Wen, Xumeng and Cao, Wei and Huang, Congrui and Bian, Jiang and Lin, Stephen and Wu, Zhirong},
journal = {Transactions on Machine Learning Research},
year = {2024},
url = {https://openreview.net/forum?id=TwiSBZ0p9u}
}

@article{goswami2024moment,
title = {{MOMENT}: A family of open time-series foundation models},
author = {Goswami, Mononito and Szafer, Konrad and Choudhry, Arjun and Cai, Yifu and Li, Shuo and Dubrawski, Artur},
journal = {arXiv preprint arXiv:2402.03885},
year = {2024}
}

@inproceedings{chemeris2026aionoscope,
  title     = {Aionoscope: Debugging Latent-State Accessibility in Time-Series Representations},
  author    = {Chemeris, Alexander and Jin, Ming and Balestriero, Randall},
  booktitle = {The 12th Mining and Learning from Time Series Workshop (MiLeTS '26), held in conjunction with KDD 2026},
  year      = {2026},
  url       = {https://github.com/langotime/aionoscope}
}

\end{document}